\documentclass{article}

\usepackage{nips_2016}

\usepackage[utf8]{inputenc} 
\usepackage[T1]{fontenc}    
\usepackage{hyperref}       
\usepackage{url}            
\usepackage{booktabs}       
\usepackage{amsfonts}       
\usepackage{nicefrac}       
\usepackage{microtype}      
\usepackage{graphicx}
\pdfoutput=1
\DeclareGraphicsExtensions{.eps,.png,.pdf}
\graphicspath{{.../images/}}

\title{Combinatorially Generated Piecewise Activation Functions}

\author{
  Justin Chen\thanks\\
  Department of Computer Science\\
  Boston University\\
  Boston, MA 02215 \\
  \texttt{chenjus@bu.edu} \\
}

\begin{document}

\maketitle

\begin{abstract}
In the neuroevolution literature, research has primarily focused on evolving the number of nodes, connections, and weights in artificial neural networks. Few attempts have been made to evolve activation functions. Research in evolving activation functions has mainly focused on evolving function parameters, and developing heterogeneous networks by selecting from a fixed pool of activation functions. This paper introduces a novel technique for evolving heterogeneous artificial neural networks through combinatorially generating piecewise activation functions to enhance expressive power. I demonstrate this technique on NeuroEvolution of Augmenting Topologies using ArcTan and Sigmoid, and show that it outperforms the original algorithm on non-Markovian double pole balancing. This technique expands the landscape of unconventional activation functions by demonstrating that they are competitive with canonical choices, and introduces a purview for further exploration of automatic model selection for artificial neural networks.
\end{abstract}

\section{Introduction}
One of the greatest challenges facing artificial neural network research is appropriate model selection. Currently, research in this field heavily relies on manually tuning hyperparameters of painstakingly contrived handcrafted architectures. With this paradigm, minor changes to an architecture means additional days or weeks to retrain that network. Furthermore, default architectures are not necessarily always the best choices or the only choices for a task. Problems could be solved more efficiently and effectively given an appropriate modality for automatic model selection. 

Previously, efforts have primarily studied {\it topology and weight evolving artificial neural networks} (TWEANNs), yet automatic activation function selection has received little attention. Of the studies done in this area, many examined heterogeneous neural networks (networks using more than one kind of activation function) by selecting from a fixed pool of hand-selected functions [10, 11, 12, 13, 16, 17, 18, 19, 20].  Most approaches stick to using conventional functions like Sigmoid and Gaussian. Few, however, have explored alternative paradigms. Yao thoroughly reviewed a variety of studies that primarily evolved network topology and activation functions [9]. Duch and Jankowski surveyed a multitude of activation functions and organized them into a taxonomy [15, 21]. 

Here, I present combinatorially generated piecewise activation functions, which is a novel approach for automatically selecting activation functions. Each generated piecewise activation function expresses, what I refer to as, a resting state, the left side, and an active state, the right side. I hypothesize that combinatorially generating activation functions enhances the expressive power of neuroevolutionary algorithms. I also demonstrate that this technique improves average accuracy and can easily be adapted into existing algorithms while only slightly increasing the average number of required evaluations and average network size, and slightly decreasing average generalization.

\section{Background}
\subsection{Neuroevolution}
Neuroevolution is a genetic programming paradigm for automatic model selection of artificial neural networks and is capable of discovering globally optimal representations for any aspect of a neural network. Neuroevolutionary algorithms represent a network as a genome, which is a collection of genes. Each gene typically represents some subset of the network e.g. a node or connection, and accompanying attributes. Training begins by generating an initial population of networks. Each generation, all genomes are evaluated on the task and assigned a fitness score. At the end of each generation, networks mate and perform crossover of genes to generate the next generation. Each component in a gene is then probabilistically mutated. After a number of generations, referred to as the drop-off age, the lowest performing portion of the population is eliminated or restricted from the rest of the population. The algorithm continues until one of the following occurs: a maximum number of generations is reached, the population collectively meets a desired fitness value, or the entire population degenerates and fails to make progress. If the population survives, a network can be selected for testing. Together, these mechanisms create artificial evolutionary pressure, search bias for an optimal genome model [14], to systematically train different architectures until they produce a globally optimal final population of networks.

Additionally, genetic algorithms do not use gradient information to optimize networks. Therefore, they are not susceptible to becoming trapped in local optima. Because of this, neuroevolution is capable of optimizing nondifferentiable functions and dynamic networks that change in size as they evolve, unlike backpropagation. Thus, neuroevolution a suitable modality for evaluating combinatorially generated piecewise activation functions. Neuroevolution techniques can also be combined with gradient optimization techniques as described by White and Ligomenides [11]. Although neuroevolution can be more attractive than gradient-based techniques, they may take more time than gradient-based methods, and may use many neuroevolutionary hyperparameters that require manual tuning before each experiment. 

\subsection{Neuroevolution of Augmenting Topologies}
{\it NeuroEvolution of Augmenting Topologies} (NEAT) is a neuroevolutionary algorithm for simultaneously evolving weights and adaptable, complex topologies for stochastic reinforcement learning control tasks. NEAT performs particularly well in continuous and high-dimensional state spaces, and outperforms many other evolutionary approaches [1]. NEAT grows networks from minimal structure. Networks start with a single input node, no hidden nodes, and a single output node, to reduce the search space, and only broadens its search as needed. Additionally, the algorithm employs a principled method of crossover to address competing conventions, when many genomes express the same phenotype (behavior). Lastly, NEAT protects innovation that may require several generations to optimize by separating genomes into species. This allows new genomes to only compete with similar genomes instead of the entire population. These three properties make NEAT ideal for evolving networks to use combinatorially generated piecewise activation functions, which is why I chose it for my experiments. Other variations of NEAT also exist and were considered [2, 3, 4], but provided unnecessary additional behaviors.

The original NEAT was tested on several pole balancing control benchmarks. Here I only focus on its hardest benchmark, double pole balancing with no velocity information. In this task, NEAT controls a virtual cart on a finite length track with two poles on top attached with hinges. It is also given a fitness score, which is calculated by the total number of times both poles remained balanced, and the angular information of the poles. Poles are considered balance as long as they do not fall more than 36 degrees from the starting position. To succeed, networks needed to balance the poles for 1,000 time steps each generation for 100 generations, which is about 30 minutes. The best performing networks from each species each generation were then tested for generalization performance. For these tests, they needed to pass at least 200 out of 625 different starting states. For each starting state, its networks needed to balance both poles for 1,000 time steps. 

\begin{figure}[ht!]
    \centering
    \includegraphics[scale=0.4]{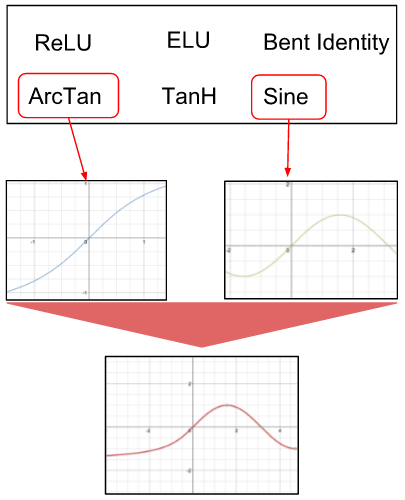}
    \caption{ArcTan and Sine Piecewise}
\end{figure}

\section{NEATwise}
For this study, I developed a modified version of NEAT, which I cleverly dubbed NEATwise. NEATwise evolves networks as the original, but when it mutates nodes into the network, it selects functions to represent the node's resting state and active state from a pool of activation functions. I was initially inspired by several previous studies [5, 6, 7, 8] that showed that ReLU, which is a piecewise activation function, performed better than others in complex architectures.

Canonical activation functions, such as the logistic function, are typically chosen for their nice mathematical properties - continuous differentiability, monotonicity, and fast computation - so networks can be optimized with backpropagation. Continuous differentiability allows for weights to be updated. Monotonicity ensures a convex error surface. Sigmoidal functions are usually bounded above and below, and are easy to compute. Although these are desired properties for networks using backpropagation, neuroevolution does not impose such constraints and is therefore capable of exploring a wider range of activation functions.

For this study, I created a pool of seven non-parametric canonical activation functions (Sine, Sigmoid, ArcTan, TanH, Bent identity, ReLU, and ELU) that could be paired together to generate piecewise smooth functions, so I could observe different mathematical properties. Each time the algorithm mutated a new node into the network, the node selected two functions uniformly at random from the pool for the resting state and the active state. This meant the algorithm could generate 49 different piecewise activation functions. For a network of {\it n} nodes, there are $49^n$ possible configurations of activation functions for a single topology. This exponential search space further justifies my choice for NEAT, which was designed to handle high-dimensional search.


\section{Observations}
After initial experiments, I discovered that NEATwise using all seven functions always failed. I observed that the population fitness continuously dropped below the fitness of the previous generation when the algorithm halted. I hypothesized that this was likely due to the exponential search space. To confirm this, I evaluated the accuracy of each canonical function on the original NEAT. The best performing networks were the out of the box NEAT which used a Sigmoid with a slope of 4.924273 and NEAT which used an unaltered ArcTan with a slope of 1 as shown in table 1. I then restricted NEATwise's pool to these two functions to reduce the search space. However, NEATwise still failed all experiments. I then observed that by increasing the drop-off age and biasing activation function selection towards ArcTan, networks survived and completed the task with a higher accuracy than the original NEAT that only used a scaled Sigmoid.  The success of the networks was due to preserving the search bias, which was being destroyed by uniform random noise.

\begin{table}[h!]
  \centering
  \caption{Out of the box and Homogeneous NEATs}
  \label{tab:table1}
  \begin{tabular}{cc}
    \toprule
    \textbf{Function} & \textbf{Success \%} \\
    \midrule
    Out of the box & 96.70 $\pm 0.016$ \\
    ArcTan            & 94.20 $\pm 0.027$ \\
    ELU 		  & 87.90 $\pm 0.045$ \\
    Sine 		  & 87.40 $\pm 0.039$ \\
    TanH 		  & 84.10 $\pm 0.028$ \\
    Bent identity   & 82.05 $\pm 0.041$ \\
    ReLU 		  & 80.00 $\pm 0.039$ \\
    Sigmoid 	  & 54.20 $\pm 0.038$ \\
    \bottomrule
  \end{tabular}
\end{table}

\section{Experiments}
My experiments demonstrate that combinatorially generating piecewise activation functions can enhance a network's expressive power. I tested NEATwise with the same {\it Double Pole No Velocities} (DPNV) parameters used by NEAT from the p2nv.ne parameter file. However, I increased the dropoff age to 50 and number of runs (experiments) to 100 after observing that the population could not optimize within 15 generations. These values were found empirically. Every 15 generations, a large portion of the population was unfit and not allowed to reproduce resulting in every experiment failing. I modified the program to gather the following statistics from all experiments in a single instance  (1 instance ran 100 experiments): total number of evaluations, total number of successful experiments, total generalization score, total number of generalization champions, total count of nodes, total sample of networks, and a sample activation function distribution taken from the final experiment of the instance. I then ran 100 instances of NEATwise for four different configurations across five 64-bit CentOS Linux 7 GNOME 3.14.2 each with an Intel Core i5-4590 CPU at 3.30 GHz x 4 and about 8 GB of RAM. Each instance used about 10\% of the CPU.

The first experiment (SA0) was used to confirm my uniform random noise hypothesis. This NEATwise generated discontinuous piecewise activation functions with an unaltered Sigmoid and unaltered ArcTan and selected each with 50\% chance. All 10,000 experiments failed as expected as shown in table 2. No statistics about the architectures could be calculated. This network introduced more nonlinearities faster than it could optimize them. This indicated that the algorithm needed to bias its selection in a way that would allow it to continue to assimilate innovative structures while optimizing existing nonlinearities at an appropriate rate.

I then studied NEATwise with an unaltered ArcTan and Sigmoid scaled by 4.924273 and shifted by -0.5, which were selected with 87.5\% and 12.5\%, respectively. Sigmoid was scaled down so that it could form continuous piecewise functions with ArcTan. I refer to this experiment at SA1. 85.60\% of experiments (88 instances) succeeded in finding solutions confirming the uniform random noise hypothesis. The bias values were found experimentally and were used for all experiments.

The third experiment (SA2) was used to determine the effects of generating discontinuous piecewise to contrast with SA1. NEATwise selected from a pool of unaltered ArcTan and Sigmoid with the out of the box slope. 81.53\% of the experiments (83 instances) succeeded in finding solution networks.

The last experiment (SA3) examined the success of unaltered Sigmoids and ArcTans. 85.60\% of the experiments (87 instances) succeeded in generating solutions. Surprisingly, this configuration performed better on average than the networks using a specially tuned slope for Sigmoid (SA2). This set of networks also showed that discontinuous functions are competitive with canonical choices. 

Although the NEATwise only succeeded at 85.60\% of experiments at most, all combinatorially generated piecewise NEAT had higher success rates than the original which only achieved 96.70\% over the course of separate 100 instances (10,000 experiments) disregarding failed experiments as shown in table 3. Typically, however, a single network is chosen from the final population of networks to use for testing, and so the difference in success rate does not reflect the value of this technique. In fact, research has been done to utilize the entire final population of neuroevolutionary algorithms [22].

\begin{table}[h!]
  \centering
  \caption{non-Markovian Double Pole Balancing Averages including failed experiments}
  \label{tab:table1}
  \begin{tabular}{ccccc}
    \toprule
    \textbf{Network } & \textbf{Success \%} & \textbf{Evaluations} & \textbf{Generalizations} & \textbf{No. Nodes}\\
    \midrule
    NEAT & 96.70 $\pm 0.016$       & 28,924.58 $\pm 37.806$       & 268.63 $\pm 0.110$       & 5.715 $\pm 0.006$ \\
    SA1 & 85.60 $\pm 0.318$       & 36,452.11 $\pm 140.073$       & 269.84 $\pm 1.008$       & 6.555 $\pm 0.025$ \\
    SA2 & 81.53 $\pm 0.371$       & 35,721.12 $\pm 167.155$       & 267.32 $\pm 1.221$       & 6.289 $\pm 0.029$ \\
    SA3 & 85.60 $\pm 0.333$       & 36,792.20 $\pm 148.148$       & 265.76 $\pm 1.038$       & 6.281 $\pm 0.025$ \\
    SA4 & 0.00 $\pm 0.00$       & 0.00 $\pm 0.00$       & 0.00 $\pm 0.00$       & 0.00 $\pm 0.00$ \\
    \bottomrule
  \end{tabular}
\end{table}

\begin{table}[h!]
  \centering
  \caption{non-Markovian Double Pole Balancing Averages excluding failed experiments}
  \label{tab:table1}
  \begin{tabular}{ccccc}
    \toprule
    \textbf{Network } & \textbf{Success \%} & \textbf{Evaluations} & \textbf{Generalizations} & \textbf{No. Nodes}\\
    \midrule
    NEAT & 96.70 $\pm 0.016$       & 28,924.58 $\pm 37.806$       & 268.63 $\pm 0.110$       & 5.715 $\pm 0.006$ \\
    SA1 & 97.31 $\pm 0.016$       & 35,763.91 $\pm 33.862$       & 269.05 $\pm 0.110$       & 6.566 $\pm 0.006$ \\
    SA2 & 98.36 $\pm 0.013$       & 35,215.19 $\pm 36.135$       & 267.97 $\pm 0.107$       & 6.28 $\pm 0.006$ \\
    SA3 & 98.37 $\pm 0.011$       & 36,453.87 $\pm 36.608$       & 266.87 $\pm 0.105$       & 6.26 $\pm 0.005$ \\
    \bottomrule
  \end{tabular}
\end{table}

\section{Future Work}
As this is a novel approach for evolving activation functions, there are plenty of opportunities for future work. One of the most challenging aspects that could be heavily improved on is developing a mechanism for more principled automatic activation function selection that adapts to the problem domain. My experiments required information about each of the canonical activation function's performance {\it a priori}. Future research could also examine the effects of different activation functions, such as parametric functions,  or distance functions as presented by Duch and Jankowski [15]. Analyzing and optimizing biases could also be further examined by defining a distribution over the pool of functions. Multi-piece piecewise activation functions could potentially allow for even greater degrees of freedom. Combinatorially generating piecewise activation functions in different architectures or with different neuroevolutionary algorithms would be very important to further explore the strengths and weaknesses of this technique. Networks generated with this technique could be examined to further delve into what suboptimal structures or mechanisms cause populations of networks to degenerate. Additionally, research could be done to determine the optimal non-topological hyperparameters such as drop-off age, mutation rate, or initial population size as the problem domain changes.

\section{Conclusions}
Combinatorially generating piecewise activation functions with NEAT outperforms the out of the box solution on the challenging stochastic reinforcement control task, DPNV. Although NEATwise failed experiments more often than NEAT, NEATwise generated networks with a higher average success rate and achieved solutions with a comparable network sizes, generalization scores, but with slightly greater number of evaluations. This technique enhances the expressive power of {\it NeuroEvolution of Augmenting Topologies}, and automatically selects appropriate models for DPNV. Giving neural networks more precise tools enables them to more accurately solve and adapt to tasks, which is important for a general purpose algorithm such as this.

\section*{References}

[1] Stanley, Kenneth O., and Risto Miikkulainen. Evolving neural networks through augmenting topologies. Evolutionary computation 10.2 (2002): 99-127.

[2] Stanley, Kenneth O. "Compositional pattern producing networks: A novel abstraction of development." Genetic programming and evolvable machines 8.2 (2007): 131-162.

[3] Stanley, Kenneth O., Bobby D. Bryant, and Risto Miikkulainen. "Evolving adaptive neural networks with and without adaptive synapses." Evolutionary Computation, 2003. CEC'03. The 2003 Congress on. Vol. 4. IEEE, 2003.

[4] Stanley, Kenneth O., David B. D'Ambrosio, and Jason Gauci. "A hypercube-based encoding for evolving large-scale neural networks." Artificial life 15.2 (2009): 185-212.

[5] LeCun, Yann, Yoshua Bengio, and Geoffrey Hinton. "Deep learning." Nature 521.7553 (2015): 436-444.

[6] Nair, Vinod, and Geoffrey E. Hinton. "Rectified linear units improve restricted boltzmann machines." Proceedings of the 27th International Conference on Machine Learning (ICML-10). 2010.

[7] Glorot, Xavier, Antoine Bordes, and Yoshua Bengio. "Deep sparse rectifier neural networks." International Conference on Artificial Intelligence and Statistics. 2011.

[8] Agostinelli, Forest, et al. "Learning activation functions to improve deep neural networks." arXiv preprint arXiv:1412.6830 (2014).

[9] Yao, Xin. "Evolving artificial neural networks." Proceedings of the IEEE 87.9 (1999): 1423-1447.

[10] Turner, Andrew James, and Julian Francis Miller. "NeuroEvolution: Evolving Heterogeneous Artificial Neural Networks." Evolutionary Intelligence 7.3 (2014): 135-154.

[11] D. White and P. Ligomenides, "GANNet: A genetic algorithm for optimizing topology and weights in neural network design," in Proc. Int. Workshop Artificial Neural Networks (IWANN'93), Lecture Notes in Computer Science, vol. 686. Berlin, Germany: Springer-Verlag, 1993, pp. 322-327.

[12] Y. Liu and X. Yao, "Evolutionary design of artificial neural networks with different nodes," in Proc. 1996 IEEE Int. Conf. Evolutionary Computation (ICEC'96), Nagoya, Japan, pp. 670-675.

[13] M. W. Hwang, J. Y. Choi, and J. Park, "Evolutionary projection neural networks," in Proc. 1997 IEEE Int. Conf. Evolutionary Computation, ICEC?97, pp. 667-671.

[14] Whitley, Darrell. "A genetic algorithm tutorial." Statistics and computing 4.2 (1994): 65-85.

[15] Duch, Wlodzislaw, and Norbert Jankowski. "Survey of neural transfer functions." Neural Computing Surveys 2.1 (1999): 163-212.

[16] Duch, Wlodzislaw, and Norbert Jankowski. "Transfer functions: hidden possibilities for better neural networks." ESANN. 2001.

[17] Weingaertner, Daniel, et al. "Hierarchical evolution of heterogeneous neural networks." Evolutionary Computation, 2002. CEC'02. Proceedings of the 2002 Congress on. Vol. 2. IEEE, 2002.

[18] Turner, Andrew James, and Julian Francis Miller. "Cartesian genetic programming encoded artificial neural networks: a comparison using three benchmarks." Proceedings of the 15th annual conference on Genetic and evolutionary computation. ACM, 2013.

[19] Sebald, Anthony V., and Kumar Chellapilla. "On making problems evolutionarily friendly part 1: evolving the most convenient representations."Evolutionary Programming VII. Springer Berlin Heidelberg, 1998.

[20] Annunziato, M., et al. "Evolving weights and transfer functions in feed forward neural networks." Proc. EUNITE2003, Oulu, Finland (2003).

[21] Duch, Wlodzislaw, and Norbert Jankowski. "New neural transfer functions."Applied Mathematics and Computer Science 7 (1997): 639-658.

[22] Yao, Xin, and Yong Liu. "Ensemble structure of evolutionary artificial neural networks." Evolutionary Computation, 1996., Proceedings of IEEE International Conference on. IEEE, 1996.

\end{document}